\documentclass[letterpaper, 10 pt, journal, twoside]{IEEEtran}
\usepackage{graphics} 
\usepackage{epsfig} 
\usepackage{mathptmx} 
\usepackage{times} 
\usepackage{amsmath} 
\usepackage{amssymb}  
\usepackage{xspace}
\usepackage{booktabs}
\usepackage{pifont}
\newcommand{\cmark}{\ding{51}}%
\newcommand{\xmark}{\ding{55}}%
\usepackage{cleveref}
\usepackage{gensymb}
\usepackage[colorlinks=true]{hyperref}
\usepackage{xcolor}

\newcommand{\New}[1]{{\color{black}{#1}}}

\newcommand*{\Cdot}{\raisebox{-0.25ex}{\scalebox{1.75}{$\cdot$}}}
\newcommand{\mypara}{\vspace*{1mm}\noindent \textbf}
\newcommand{\myitem}{\vspace*{0mm}\item[\Cdot]}

\DeclareMathOperator*{\argmax}{argmax}
\def\eg{e.g., }
\def\ie{i.e., }

\def\etal{et al. }
\begin{document}
%
\title{Grasping in the Wild: Learning 6DoF Closed-Loop Grasping from Low-Cost Demonstrations}
\author{
\vspace{-2mm}
Shuran Song$^{1, 2}$\quad\quad
Andy Zeng$^{2}$\quad\quad
Johnny Lee$^{2}$ \quad\quad
Thomas Funkhouser$^{2}$
\vspace{3mm}
\\
\href{https://graspinwild.cs.columbia.edu}{https://graspinwild.cs.columbia.edu}\quad
\thanks{$^{1}$ Columbia University {\tt\footnotesize shurans@cs.columbia.edu}} 
\thanks{$^{2}$ Google {\tt\footnotesize andyzeng@google.com, johnnylee@google.com, tfunkhouser@google.com}}%
\thanks{Digital Object Identifier (DOI): see top of this page.}
}

\markboth{IEEE Robotics and Automation Letters. Preprint Version. Accepted June, 2020}
{Song \MakeLowercase{\textit{et al.}}: Grasping in the Wild}

\maketitle

\begin{abstract}
Intelligent manipulation benefits from the capacity to flexibly control an end-effector with high degrees of freedom (DoF) and dynamically react to the environment.
However, due to the challenges of collecting effective training data and learning efficiently, most grasping algorithms today are limited to top-down movements and open-loop execution. 
In this work, we propose a new low-cost hardware interface for collecting grasping demonstrations by people in diverse environments.
This data makes it possible to train a robust end-to-end 6DoF closed-loop grasping model with reinforcement learning that transfers to real robots.
A key aspect of our grasping model is that it uses ``action-view'' based rendering to simulate future states with respect to different possible actions.
By evaluating these states using a learned value function (e.g., Q-function), our method is able to better select corresponding actions that maximize total rewards (i.e., grasping success).
Our final grasping system is able to achieve reliable 6DoF closed-loop grasping of novel objects across various scene configurations, as well as in dynamic scenes with moving objects. 
\end{abstract}


%
\IEEEpeerreviewmaketitle

\section{Introduction}
Versatile manipulation benefits from the capacity to flexibly control an end-effector in 3D space and dynamically react to changes in the environment.
In the case of grasping, 6 degrees of freedom (6DoF: where the gripper is free to change in x, y, z position and in roll, pitch, yaw) closed-loop algorithms enable robots to pick up objects from a wider range of unstructured settings beyond tabletop scenarios: from moving in 6DoF to retrieve diagonally positioned plates in a dishwasher or harvest berries from a bush, to using closed-loop visual feedback for grasping objects moving along a conveyor belt or handed off by people.
Despite the practical value of both 6DoF control and closed-loop feedback, most data-driven grasping algorithms today are only able to achieve one of these capabilities.
Most methods only infer top-down grasps (4Dof: x, y, z, yaw) in simple tabletop settings \cite{pinto2016supersizing, zeng2018robotic,zeng2018learning,yen2020learning}, or detect grasps in 6DoF but with open-loop execution \cite{lu2018planning,mousavian20196}.

\begin{figure} [t]
    \centering
    \includegraphics[width=\linewidth]{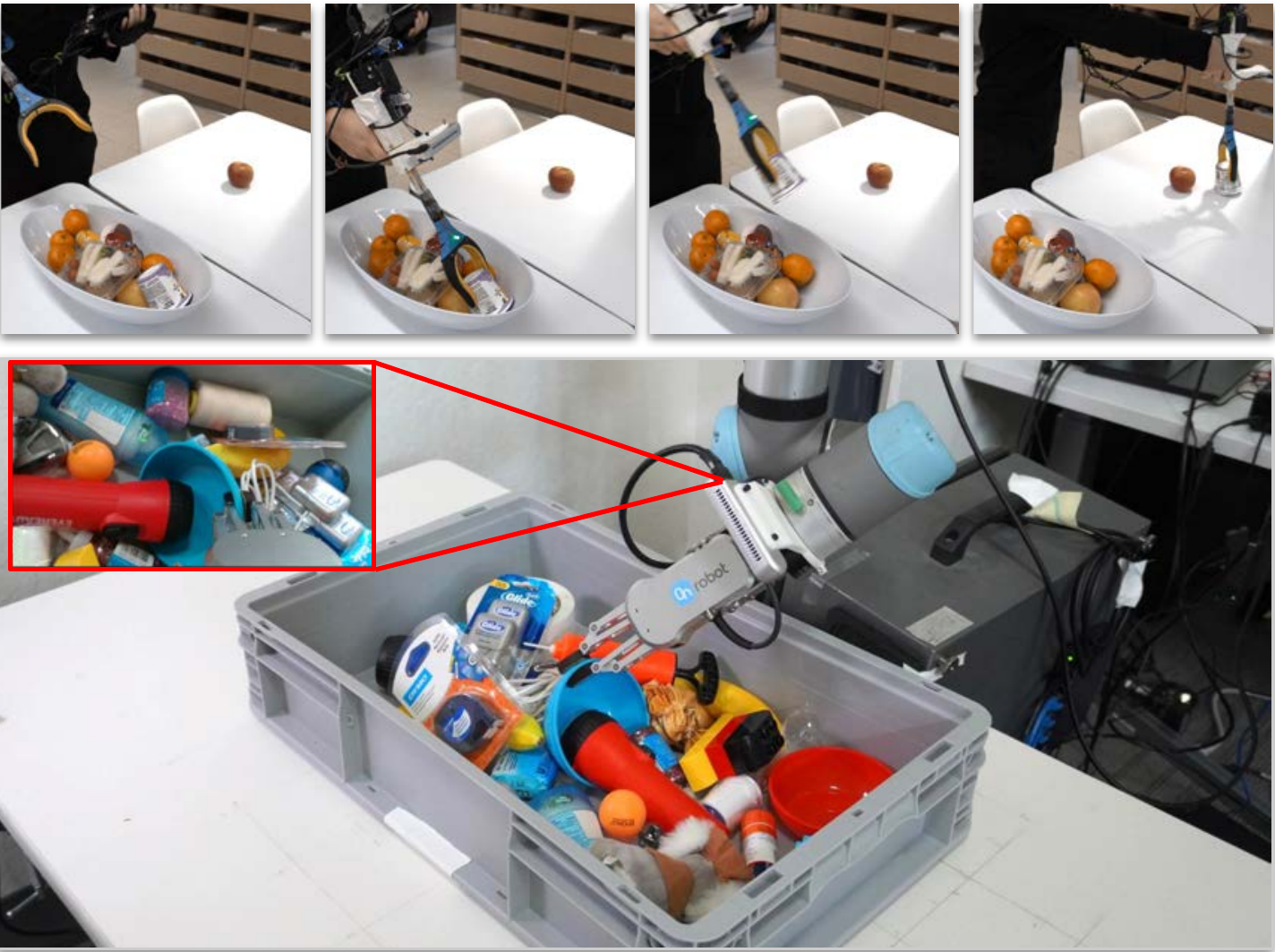}
    \caption{\textbf{Grasping in the wild.} We developed a low-cost handheld device that enables people to collect grasping demonstrations (top row) while carrying out everyday tasks in diverse environments. Using these demonstrations as training data, we show that it is possible to learn flexible 6DoF closed-loop grasping policies that transfer to real-world robot picking systems (bottom).
    }
    \label{fig:teaser}
    \vspace{-3mm}
\end{figure}

One major obstacle for achieving both 6DoF and closed-loop grasping is the challenge of acquiring effective training data.
Collecting data on real robots through self-supervised trial and error is expensive. As the action space approaches higher dimensions (\eg 4DoF to 6DoF grasping) and as the state space reaches higher diversity (\eg images of static scenes to dynamic scenes), the exploration search space grows exponentially. In this large search space, the chances of stumbling on useful grasping trajectories through random search becomes exponentially slim. 
While prior work alleviates some of these issues by training on demonstration data collected from human teleoperation of robots \cite{zhang2018deep}, these approaches remain limited to a small range of environments that are physically accessible for those robots. 

In this work, we develop a system for collecting grasping demonstrations in the wild by equipping a handheld grabbing tool with an RGB-D camera mounted on its ``wrist'' in the same way it would be on a real robot arm (Fig.~\ref{fig:teaser}).
This device (which in total costs \$600) is a low-user-friction tool that can be used by people to pick up objects while carrying out everyday tasks real-world environments (\eg picking up trash, sorting dishes, etc.).
During these tasks, the camera captures RGB-D gripper-centric videos from which we recover 6DoF grasping trajectories using classic visual tracking algorithms.
This setup provides grasping demonstration data with substantially higher diversity and lower cost than prior work.

This data makes it possible to bootstrap and train a robust end-to-end 6DoF closed-loop grasping model with reinforcement learning that transfers to real robot platforms.
The system uses a deep network to model a value function that maps from a visual observation of the state (\ie gripper-centric images) to the expected rewards in that state.
A key aspect of our grasping model is that it uses \emph{``action-view''} based rendering to simulate future states with respect to different possible actions (\eg what the gripper camera would see if it moves forward or sideways).  It evaluates these states using the learned value function in a closed-loop while executing grasps to predict how the gripper should move in the next time-step to maximize rewards.  


In summary, our main contributions are 1) a real-world dataset of human grasping demonstrations in diverse environments collected using a new low-cost hardware interface, and 2) a visual 6DoF closed-loop grasping algorithm that uses action-view based rendering to achieve 92\%  grasping success rates in static scenes and 88\% in dynamic scenes with moving objects. Our experiments demonstrate that the capacity to move in 6DoF enables our system to grasp novel objects in a variety of environments: from grasping objects sideways from a wall to picking from inclined bins. We also show that the performance and learning efficiency substantially improves by training on demonstration data collected with our tool. Qualitative results are available in our supplemental video at \href{https://graspinwild.cs.columbia.edu}{https://graspinwild.cs.columbia.edu}


\section{Related Work}

In this section, we review relevant work on vision-based grasping and data collection for data-driven grasping. 

\mypara{Vision-based grasping.}
Classic vision-based grasping solutions often explicitly model contact forces with prior knowledge of object geometry, pose, and dynamics \cite{prattichizzo2008grasping, weisz2012pose, goldfeder2009columbia, zeng2017multi}. 
However, this kind of prior knowledge is difficult to obtain for novel objects in unstructured environments.

More recent data-driven methods explore the prospects of training object-agnostic grasping policies that detect grasps by exploiting learned visual features, without explicitly using object-specific knowledge \cite{redmon2015real,pinto2016supersizing,pinto2017learning,gualtieri2016high,mahler2017dex,zeng2018robotic, lu2018planning,mousavian20196,gualtieri2018learning}.
This problem formulation enables these methods to generalize to novel objects without the need for scanning the objects to obtain 3D models or estimate their poses. 
However, since most of these approaches perform open-loop grasp execution, they are sensitive to calibration errors and fail to handle dynamic environments.

Another line of work tackles closed-loop grasping by designing algorithms that continuously gather visual observations during grasp execution and predict next actions using visual servoing \cite{viereck2017learning,morrison2018closing} or reinforcement learning \cite{kalashnikov2018qt}. 
However, these methods are characterized by constrained state-action spaces in order to reduce the amount of training data required. 
For example, QT-Opt \cite{kalashnikov2018qt} learns only top-down grasping policies (action space) with images from a fixed static camera (state space). As a result, the system cannot immediately generalize to different task configurations (\eg grasping from shelves) without extensive retraining. Specifically, QT-Opt trains using a total of 580k off-policy + 28k on-policy grasping trials to learn an effective policy for the current setup, which makes it challenging to generalize to larger state-action spaces.  
In this work, we propose to use human demonstration and action-view representations to improve learning efficiency.

\begin{table}[h]
\centering
\caption{\textbf{Comparisons of visual grasping algorithms.}} 
\vspace{-3mm}
\setlength\tabcolsep{4px}
\begin{tabular}{l ccl  }
\toprule
Method & Closed-Loop & 6DoF &  Training Data  \\ \midrule
\cite{mahler2017dex,mahler2016dex} &\xmark& \xmark & simulation \\
\cite{pinto2016supersizing, zeng2018robotic,zeng2018learning,levine2018learning,lenz2015deep}     &\xmark& \xmark &  real \\
\cite{gualtieri2016high,lu2018planning,mousavian20196,murali20196, gualtieri2018learning}     &\xmark& \cmark & simulation \\
\cite{viereck2017learning} &\cmark& \xmark &  simulation  \\
\cite{morrison2018closing,kalashnikov2018qt} &\cmark& \xmark &  real  \\
Ours                       &\cmark &\cmark & real \\ 
\bottomrule
\end{tabular}
\end{table}

\begin{figure*}[t]
    \centering
    \includegraphics[width=\linewidth]{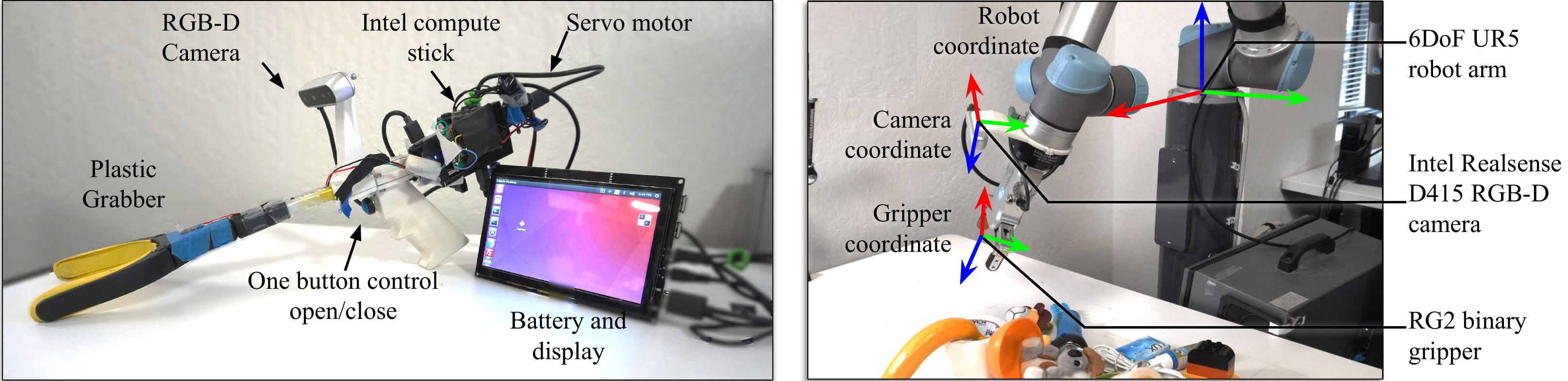}~ 
    \caption{\label{fig:gripper} \textbf{Hardware setup.} Our low-cost handheld device (left) consists of a plastic grabber tool equipped with an RGB-D camera and a servo that controls the binary opening of the grabber fingers. This device was designed to be analogous to the real robot's end effector setup (right), while providing a low-user-friction interface that enables untrained people to collect grasping data in almost any environment.
    }
    \vspace{-3mm}
\end{figure*}

\mypara{Grasping data acquisition.}
Learning-based grasping algorithms heavily depend on acquiring high-quality training data.
However, most prior self-supervised grasping systems are often constrained to learning in simulation \cite{viereck2017learning,mahler2017dex,mahler2016dex,sajjan2019cleargrasp} or structured lab environments \cite{pinto2016supersizing, zeng2018learning,levine2018learning,lenz2015deep}. Gupta \etal \cite{gupta2018robot} improves the data collection process by physically moving a robot into different environments. However, the data is still limited to simple scenarios (\eg picking up toys from the floor) due to inefficient exploration algorithms (with low initial grasping success rates) and constrained physical robot access to diverse environments. 

Learning from demonstration is a popular approach to address sample efficiency problems.  With human experts directly annotating the training data \cite{zeng2018robotic,lenz2015deep} or controlling the robot via teleoperation \cite{zhang2018deep,sharma2018multiple}, the system can quickly obtain positive examples to speed up the training process. 
However, both settings (annotation or teleoperation) require human experts to be  familiar with the robot hardware and grasping mechanisms in order to correctly annotate the grasp poses or successfully teleoperate the robot. 
Training human experts for such tasks can be expensive and difficult to scale. 
On the other hand, recording videos of direct interactions between human hands and objects does not require expert knowledge from the subject \cite{Chang2014,amor2012generalization}. However, there is often a big domain gap between the kinematics between the human hand and the robot gripper, which makes it challenging to learn transferable knowledge to robot manipulation policies. 
Praveena et al. \cite{praveena2019characterizing} also developed a similar handheld grabber tool as our setup, where the tool is equipped with force-torque sensor, and its movement is tracked with a Optitrack motion capture system. While this device allow the human user to easily collect high quality demonstration data, the setup is limited to lab environments with the motion capture system. 
In this paper, by using RGB-D reconstruction, our data collection process is designed to be accessible to inexperienced users, scalable to any environment, applicable to any task, and transferable to real robot manipulation.


\section{Approach}

Our goal is to achieve reliable 6DoF closed-loop grasping in a framework that is flexible enough to handle novel objects and dynamic scene configurations with moving objects.
We show that this goal is achievable by training visual grasping value functions (using view-based rendering for data augmentation) on a large dataset of human demonstrations (collected from a handheld gripper equipped with a wrist-mounted camera).
Sec.~\ref{sec:data} describes our hardware setup and data collection process for gathering human grasping demonstrations from a diverse set of tasks and environments (\ie in-the-wild).
Sec.~\ref{sec:alg} describes our 6DoF closed-loop grasping model and how it is trained with this data.

\section{Grasping Demonstrations In-the-Wild}
\label{sec:data}
To collect grasping data from human demonstrations, we built a low-cost portable handheld grabber tool equipped with a wrist-mounted RGB-D camera (illustrated in Fig.~\ref{fig:gripper}).
We then asked willing participants to use the tool in place of their hands for everyday pick-and-place tasks, \eg picking items from shelves, bins, refrigerators, sorting dishes in a dishwasher, or picking trash on the floor, etc. 
Our data collection system is driven by 3 key motivations:

\begin{itemize}[]
\myitem \textbf{Accessibility for diversity.} Our handheld tool is a low-user-friction interface that allows untrained people to collect manipulation data in almost any environment (\eg various homes, offices, warehouses, grocery stores), many of which would otherwise be difficult for robots to acquire physical access to. This substantially improves the diversity of the data that we can acquire.
\myitem  \textbf{Data for challenging tasks.} For challenging manipulation tasks like searching for dishes in a dishwasher, data collection through robot trial and error can be expensive -- robot failures may lead to negative irreversible consequences (\eg broken dishes). In contrast, our setup enables skilled humans to easily collect manipulation data for these tasks with negligible failure rates.
\myitem \textbf{Minimized domain gap.} Our gripper tool is designed to be as similar as possible to a real robot's end effector: binary actuated parallel-jaw fingers with a wrist-mounted RGB-D camera.
This similarity narrows the domain gap between the data collected from human demonstrations and the data that the robot encounters.
\end{itemize}


\begin{figure*}[t]
    \centering
    \includegraphics[width=\linewidth]{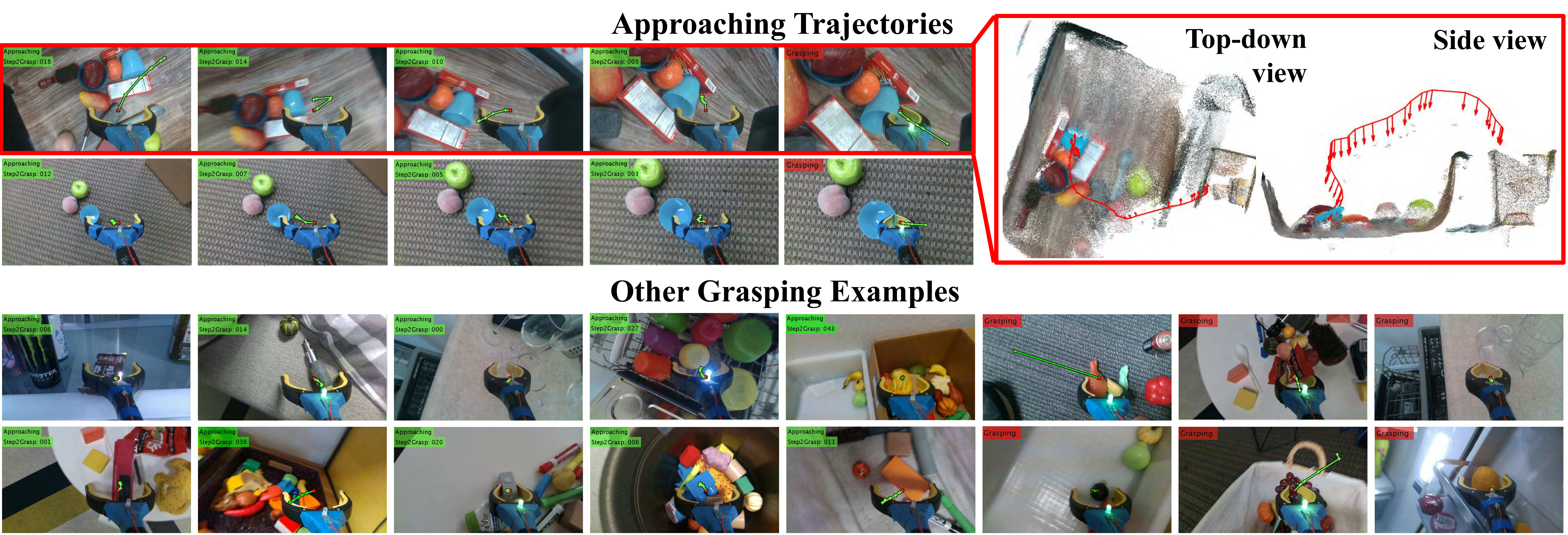}
    \caption{\textbf{Diverse demonstrations.} As the handheld device approaches a target object (\eg blue cup), RGB-D video frames (first row) are used to recover the 6DoF motion trajectory and reconstruct a 3D representation of the scene (top right). Grasping trajectories for the same object (\eg blue cup, second row) can vary depending on the object's pose in the scene, the environment, or the device user. Overall, our grasping dataset contains grasping demonstrations with a diverse set of objects, tasks, and environments (examples, bottom two rows).
    }
    \label{fig:data}
    \vspace{-3mm}
\end{figure*}

\subsection{Hardware Setup}

Our handheld data collection device (Fig.~\ref{fig:gripper}) consists of: 1) a Royal Medical Solutions (RMS) plastic grabber reacher tool forearm, 2) a Dynamixel servo that twists the grabber's internal cable to control the opening of the fingers, 3) a 3D printed grip that attaches to the back end of the grabber, 4) a binary push button on the grip that connects to an Arduino to trigger the Dynamixel servo, 5) an Intel RealSense D415 camera mounted 25cm from the gripper fingertips, streaming $640\times480$ RGB-D images to 6) an Intel compute stick running Linux OS with data capturing software, 7) a portable 12V battery to power the tool for 5 hours on a single charge, and 8) an optional touch screen monitor. All components are either purchased off-the-shelf or 3D printed with PLA. The cost of the entire unit sums to around \$600.

We designed the handheld gripper to be analogous to the end effector of the real robot setup (shown in Fig.~\ref{fig:gripper} Right), which consists of a 6DoF UR5 robot arm with an binary RG2 gripper, and an wrist-mounted Intel RealSense D415 camera. The handheld gripper uses binary control (triggered by the push button) to mimic the RG2's binary open/close behavior.



\subsection{Data Collection and Processing}

We distributed data collection among 8 participants, who were tasked with collecting grasping data while performing various pick-and-place tasks (\eg picking from shelves, picking from bins, rearranging objects, picking up trash, etc.) in different environments (\eg apartments, kitchens, offices, warehouses). The varying tasks and environments naturally encourage human demonstrators to perform different grasping strategies, which subsequently lead to more diverse demonstration data. Our dataset in total contains 12 hours of recorded gripper-centric RGB-D videos, labeled with the binary signal of when the user pushed the button to close the gripper.

To recover 6DoF grasping trajectories from the RGB-D videos of demonstrations, we use classic frame-to-frame visual tracking \cite{xiao2013sun3d} to estimate the camera pose and trajectory over time. Since the camera is fixed on the gripper and the rigid transform between the camera and gripper is calibrated and known beforehand, this tracking process also enables us to recover the gripper pose and trajectory over time. Specifically, to estimate the relative pose transform between two RGB-D frames, we detect SIFT keypoints \cite{lowe2004distinctive} on both frames and use random sample consensus (RANSAC) on correspondences, with singular value decomposition (SVD) to compute a rigid transform. We then refine that estimate by using iterative closest point (ICP) \cite{besl1992method} on the 3D point clouds projected from the frames. This algorithm makes the assumption that the environment is static -- hence to reduce noisy estimates, we mask out the pixels that belong to the gripper and grasped objects.

Additionally, we split the RGB-D videos into short clips that correspond to each picking attempt by using a set of heuristics on the binary gripper closing signal. The frames that occur before a button push (to close handheld gripper fingers) record the pre-grasp trajectory, while the frames that occur between the button push and the following button release record the post-grasp trajectory. We can also recover and track the pixel mask of the target object by using background subtraction to detect pixel regions in the images that are stationary throughout the frames captured between button push and release.

In summary, we extract the following information from each RGB-D video segment corresponding to each picking attempt: 1) pre-grasp gripper trajectory, 2) final gripper grasping pose, 3) target object pixel mask, 4) post-grasp (placing) gripper trajectory, 5) and picking order. In total, the dataset contains 7,797 valid picking attempts and grasping trajectories. Fig.~\ref{fig:data} illustrates several example demonstrations in the dataset and the grasping trajectories.

\begin{figure*}[t]
    \centering
    \includegraphics[width=\linewidth]{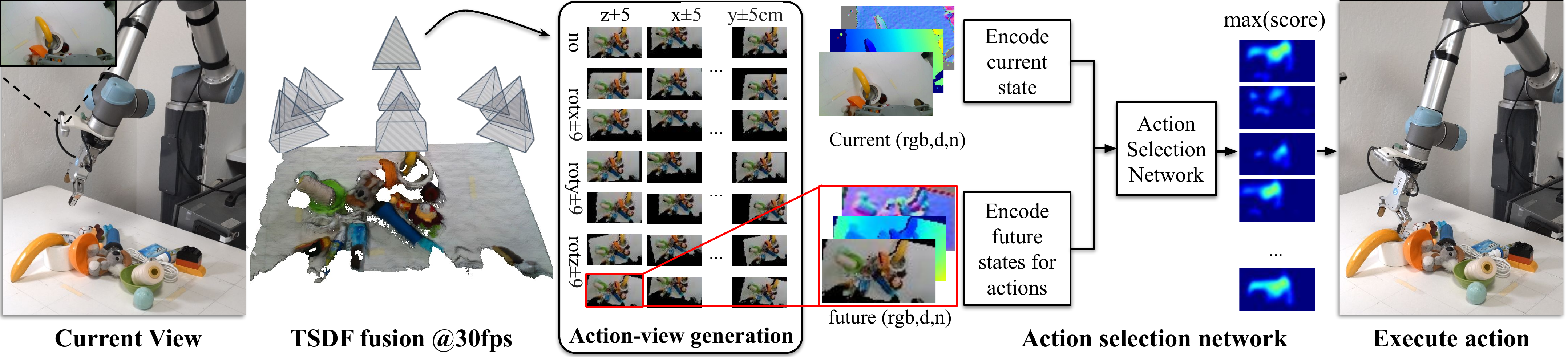}
    \caption{\textbf{Action-view based grasping overview.} From left to right, the images show: 1) current camera observation, 2) 3D scene representation from TSDF fusion 3) generated action-view pairs using view-based rendering, and 4) action-view selection network that predicts dense Q-values for each action-view pair.
    The action-view rendering step allows the algorithm to forward-simulate the set of possible future states conditioned on the current state and action. This formulation improves learning efficiency by removing the need to learn to interpret how an action should correspond to changes in the state space.
    }
    \label{fig:network}
    \vspace{-3mm}
\end{figure*}
\section{6DoF Closed-loop Vision-based Grasping \label{sec:alg}}

The task of closed-loop grasping requires an action policy that enables the robot to move its gripper towards an object, approach it from an angle that is likely to lead to a stable grasp. 
This pre-grasp approaching process is a time-varying sequence of actions, for which rewards are loosely defined, and has previously been shown to be more effectively learned through reinforcement than from direct supervision \cite{zeng2018learning,kalashnikov2018qt}.

We formulate this vision-based grasping problem as a Markov decision process: given state $s_t$ at time $t$, the robot chooses and executes an action $a_t$ according to a policy $\pi(s_t)$, then transitions to a new state $s_{t+1}$ and receives a reward $r_t$.
The goal of reinforcement learning is to find an optimal policy $\pi^*$ that selects actions which maximize the total expected rewards $Q(s_t,a_t)=\sum_{i=t}^{\infty}\lambda^{i-t} r_i$, \ie $\lambda $-discounted sum over an infinite-horizon of future returns from time $t$ to $\infty$. In this work, we use off-policy Q-learning to learn the optimal parameterized Q-function $Q_\theta(s_t,a_t)$ (\ie state-action value function), where $\theta$ might denote weights of a neural network. Formally, our learning objective is to iteratively minimize the temporal difference error $\delta_t$ between $Q_\theta(s_t,a_t)$ and a target value $y_t$: 
\begin{equation}
    \delta_t = |Q_\theta(s_t,a_t) - y_t|
    \label{eq:loss}
\end{equation}
\vspace{-4mm}
\begin{equation}
y_t=r_t+\lambda\,Q_\theta(s_{t+1},\argmax_{a_{t+1}}(Q_\theta(s_{t+1},\mathcal{A}_{t+1})))
\vspace{-1mm}
\end{equation}
\noindent where $\mathcal{A}_t$ is the set of all available actions at time $t$.

Within our formulation, we represent each state $s_t$ as a image observation from the wrist-mounted camera.
We parameterize each action $a_t$ as a 6DoF rigid transform that encodes the relative rotation and translation from the current robot end effector pose to the next target pose.
Motion planning between end effector poses is autonomously executed on the real robot using standard proportional-derivative (PD) control with inverse kinematics (IK) solvers.
The algorithm outputs a gripper closing signal by using depth observations from the camera to measure proximity to objects. The algorithm checks the local region of depth values between fingertips, and issues a close command if the nearest 1\% of depth in this area is smaller than a $d_mathrm{close}$ = depth of fingertips - 0.015m. After the gripper attempts to close, the system lifts the gripper up 0.1m and checks the finger width to determine grasp success.
Each grasping trajectory begins with the end effector initially positioned 50cm away overlooking the scene of objects, and terminates after 40 state transitions or after a successful grasp.
Rewards are provided $r_t=1$ for successful grasps and $r_t=0$ otherwise.

\subsection{View-based Rendering as Predictive Models}

The key aspect of our formulation is that at each time step $t$, we use view-based rendering to forward-simulate the set of possible future states $\hat{\mathcal{S}}_{t+1}$  conditioned on the current state $s_t$ and action taken $a_t\in{\mathcal{A}_t}$.
In other words, view-based rendering is used as a predictive model $f(s_t,a_t)=\hat{s}_{t+1}\in{\hat{\mathcal{S}}_{t+1}}$ where $\hat{s}_{t+1}$ approximates $s_{t+1}$.
Since states $\mathcal{S}_t$ are represent by wrist-mounted camera views, and possible actions $\mathcal{A}_t$ represent relative 6DoF rigid transforms of the end effector from its current pose, forward-simulating future states $f(s_t,a_t)=\hat{s}_{t+1}$ amounts to rendering a new camera view as if the end effector had moved according to $a_t$. The views are rendered with a smaller resolution ($45\times80$) to speed up both rendering and inference time.
We train our Q-functionfrom human demonstration data and fine-tune with real world trial and error (Sec. \ref{sec:alg-train}). During test time, at any given state $s_t$, our system evaluates state-action pairs using trained Q-function $Q_\theta(f(s_t,\mathcal{A}_{t}),\mathcal{A}_{t})$, and executes the action that maximizes the predicted Q-values \ie $\argmax_{a_t}(Q_\theta(f(s_t,\mathcal{A}_{t}),\mathcal{A}_{t}))$.

\New{This action-view representation is inspired by prior work, which use predictive models to improve the sample efficiency of reinforcement learning algorithms \cite{ebert2018visual,xu2019densephysnet}. In this work we show that view-based rendering with 3D reconstructions can serve as a strong proxy for predictive models in ego-centric visual grasping. In contrast to abstract action representations such as end effector Cartesian offsets or joint angles, where the mapping between the action space and state space needs to be explicitly learned (or in many cases, memorized) by the network, our action representation representation improves learning efficiency by directly representing each action (\eg gripper movement) with its corresponding future state. 
}



The grasping algorithm consists of three components: 1) a 3D reconstruction pipeline that accumulates camera observations over time to generate 3D representation of the scene, 2) a method for quickly rendering 3D scenes from arbitrary viewpoints, and 3) a deep neural network that models the value function $Q_{\theta}$. The following paragraphs describe the details of these components:

\mypara{Aggregating visual observations.} 
As the end effector approaches a target object, the wrist-mounted camera continually gathers new RGB-D images of the scene.
Due to object occlusions and clutter, each observation is partial, hence the system requires an algorithm that can aggregate these partial observations into a complete 3D scene representation.
Meanwhile, the representation should continually update itself with new observations to handle dynamic environments.

To this end, we use the Truncated Signed Distance Function (TSDF) representation for fusing observations into a 3D voxel grid, where each voxel stores a value that represents its distance to the closest surface.
The sign of that value indicates whether the voxel is in free space or occluded space \cite{curless1996volumetric, newcombe2011kinectfusion, zeng20163dmatch}. 
Our implementation stores the color of surface as well, to support ray casting for downstream view-based rendering.
At the beginning of each grasping attempt (episode), our system initializes a 3D voxel grid in robot coordinates, with voxel size set to 5mm. 
Given each new  observation \New{(\ie $360\times640$  RGB-D image) } and camera extrinsics, the system transforms the observed surface from camera coordinates into TSDF voxel grid coordinates, and updates the TSDF values for all observed voxels respectively using an exponential moving average with $\alpha=0.8$ that biases towards new observations. \New{The camera extrinsics are obtained by using robot end effector poses and a calibrated transformation between the camera and end-effector. Our UR5 robot arm features industrial-grade sub-millimeter repeatability, which enables accurate end effector poses to provide high quality reconstructions.} The region that is not directly observed by the camera (missing depth, occluded, or outside camera FoV) will remain unchanged. 

In this way, the algorithm is not only able to build a more complete 3D representation of a static scene by aggregating past observations, but is also update the representation for dynamic environments with new observations. 
Compared to other methods of aggregating past observations such as using recurrent neural networks or LSTMs \cite{hochreiter1997long}, our TSDF fusion explicitly leverages accurate industrial-grade robot motion in order to reduce the burden of learning view point registration or 3D reconstruction inside the network.

\mypara{Generating action-views.}
%
%
%
At each time step $t$, our formulation chooses between a set of $n$ ($n=35$ in our experiments) possible action candidates $a^i_t(\phi,\tau)\in\mathcal{A}_t$ each action encodes the relative rotation $\phi$ and translation $\tau$ between the current end effector pose and the next target pose.
\New{The 35 candidate actions are heuristically generated using combinatorial transforms with 5 translations x = +/-d, y=+/-d,  z=d, and 7 rotations $R_x =\pm a$, $R_y =\pm a$, $R_z =a$, and $R=0$, where $d = 0.015 + ratio*0.035$,  $a = 10+ ratio*20$, and $ratio=max\{0,min\{1, MED(D)-0.1)/0.4\}\}$ and ${MED(D)}$ is median depth value from the camera. All actions have a small z-offset of 0.01m to encourage the gripper to move forward.  }
Actions that cause self-collision or move outside the workspace are automatically removed.


%

By ray-casting the TSDF of the scene, we render virtual observations $\hat{s}^i_{t+1}$ of the robots' camera as if it had moved accordingly to action $a_{t}^i$. 
$\hat{s}^i_{t+1}$ contains an RGB-D and an surface normal image. 
After that, all generated views $\{\hat{s}^0_t ...  \hat{s}^n_t\}$ are fed into the Q-function. The state-action pair with the highest Q-value is selected and executed on the robot.

\mypara{Evaluating action-views.}
Given a set of candidate views $\{\hat{s}^0_t ... \hat{s}^n_t\}$, the goal of the network is to evaluate the Q-value with respect to each candidate and select the best corresponding action $a_t^i$ to perform.
We model our Q-function $Q_\theta(s_t,a_t)$ as a feed-forward fully convolutional network that has two input branches and one output branch. One input branch takes as input the visual observation of the state $s_t$, the other branch takes in the candidate views $\hat{s}_{t+1}^i$. The encoded current and future state features are then concatenated and fed into the action selection network to output a dense pixel-wise map of Q-values with the same image size and resolution as that of $s_t$.  
Both the state encoder and action selection networks are modeled by ResNet-18 network architectures. 
The training objective is to minimize the error $\delta_t$ between the predicted and target Q-values.
%
Section \ref{sec:alg-train} provides more details on how the target Q-value $y_t$ is assigned. 




\subsection{Learning from Human Demonstrations \label{sec:alg-train}}

Our system bootstraps its learning of the value function $Q_\theta$ from our human demonstration data.
While human demonstrations provide a diverse set of examples for learning grasping strategies, 
there are still two major issues that need to be addressed in order to make these demonstrations an effective data source for training robot grasping algorithms: 
1) like most learning from demonstration datasets, the training data distribution is naturally unbalanced: it consists of mostly positive examples, with very few negative examples.  
2) despite efforts on making the hardware setup similar, there is still a small domain gap between the demonstration data and real robot data. 
We address the first issue through negative trajectories synthesis,  
and tackle the second issue by fine-tuning on the real robot using trial and error. 

\mypara{Synthesizing negative trajectories via rendering.}
Each successful grasping demonstration trajectory (\ie episode) consists of a sequence of RGB-D images captured up until the gripper closing signal that terminates the episode. Each RGB-D image is associated with a 6DoF camera pose computed from RGB-D visual tracking (described in Sec. \ref{sec:data}). At each time step $t$ of the sequence, we use TSDF fusion to aggregate camera observations up until the current frame, then use view-based rendering with the fused volume to generate a set of action-views $a_t \in \mathcal{A}_t $ around the current camera pose (in the same fashion as our algorithm described in Sec. \ref{sec:alg}). 
\New{All action-views are ranked by their distances to the ground truth (measured by the IoU of the 3D view frustum between the candidate and ground truth view). We treat the first view as positive, other views ranked lower than the top 4 are considered as negatives. }
To balance training, we randomly sample negative views to maintain a 1:5 positive to negative example ratio.
%
The target $y_t$ value of positive views are assigned as $y_t(s_t,a_t^{pos})=\lambda^{(m-t)}$, where $t$ is number of steps in this grasping attempt, $m$ is the total step length of the grasping episode, and our discount factor $\lambda=0.999$. The $y_t$ value for all negative actions are assigned as $y_t(s_t, a_t^{neg})=0$. Note that this labeling scheme is strictly only for bootstrapping (\ie pretraining) our Q-function from demonstrations with supervised learning (while ensuring that the network satisfies the Bellman equation). This is similar to the n-step Q-learning loss for learning from demonstrations in \cite{hester2018deep}, but simplified since our rewards are sparse and only imparted at the end of each trajectory based on final grasp success.
Additionally, rather than predicting one Q-value per image, we predict pixel-wise Q-values where supervision is provided to the pixel of the final grasping pose (\ie 3D gripper position) back-projected onto the current action-view image. 
\New{The issue with predicting a global Q-value for the entire image was that after reducing the feature map into a single prediction value (\eg via max-pooling) the model tends to predict similar values for different rendered views and struggles to converge in training. We conjecture that it is because local visual and geometric details (which provide important information for grasping) are easily lost through max-pooling operations. Predicting dense Q-values for every pixel forces the network to focus on local geometric features, by specifically backpropagating gradients on local visual features that contribute most to its Q-value.}




\mypara{Fine-tuning with robot trial and error.}
To address the domain gap between data collected from human demonstrations and data from the real robot, we further fine-tune our grasping models on the real robot platform through trial and error.
During fine-tuning, our formulation trains with standard off-policy Q-learning, where target values are predicted Q-values of the next state, and no loss is backpropagated for actions not taken.
The robot executes grasping trajectories that follow the action-view Q-function predictions (pretrained from human demonstrations) with $\epsilon$-greedy exploration, where $\epsilon$ is initially fixed at 0.1, then annealed over time. 
This exploration step enables the algorithm to explore other possible grasping trajectories beyond what it has learned from demonstrations. 
After each grasping attempt (\ie episode), the new observations, action trajectories, and final binary grasping label (success or failure) are stored into the replay buffer for fine-tuning.
%
Both models with and without this fine-tuning step are evaluated.


\section{Experiments}
In this section we evaluate the effectiveness of our proposed algorithm as well as its ability to adapt to different test environment settings.
The experiments in Tab. \ref{tab:setup} and \ref{tab:dynamic} are tested on novel objects. The evaluation metric is the grasping success rate: $ = \frac{ \text{\# successful grasps}}{\text{ \# grasping episodes}}$.
\New{Table \ref{tab:dynamic} reports algorithmic run times. On average, our algorithm takes 0.18s in total for each action step using an Nvidia GPU GTX 2080Ti. The reconstruction runs at 30 FPS asynchronously with the grasping model, and does not block action execution. The rendering pipeline (with GPU parallelization) raycasts into the current TSDF volume to generate an action-view. The rendering takes 0.057s in total for all views, which are passed to the view selection network as a batch.}

\begin{table}[h!]
\vspace{-1mm}
\centering
\caption{Testing on different scene configurations (mean \%).\label{tab:setup}}

\setlength\tabcolsep{15px}
\begin{tabular}{@{}lcccc@{}}
\toprule
& Tabletop & Bin  & Wall & Random \\ \midrule
pretrain only& 76  & 66 & 78  &62 \\
+finetune    & 92   & 82  & 89   &76  \\ \bottomrule
\end{tabular}
\end{table}
\begin{figure}[h!]
\vspace{-2mm}
\centering
\includegraphics[width=\linewidth]{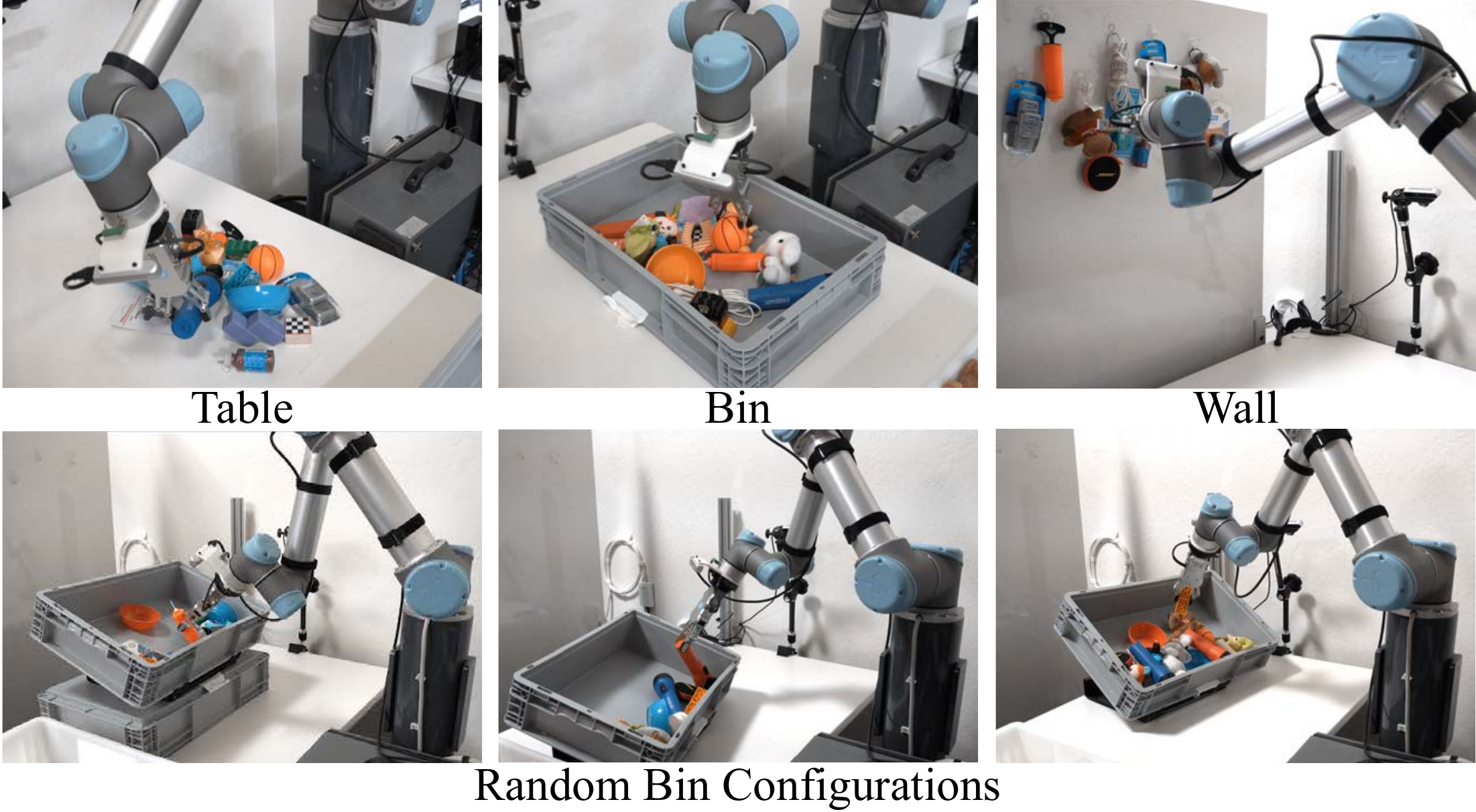}
\label{fig:diffpose}
\vspace{-5mm}
\end{figure}

\mypara{Grasping in various static settings.} 
We first investigate our algorithm's grasping performance across various static environment settings and scene configurations:
\begin{itemize}[]
    \myitem \textit{Tabletop.} Robot grasps from a pile of objects randomly dumped on a flat tabletop.
    \myitem \textit{Bin.} Robot grasps from a pile of objects randomly dumped into a bin. This is more challenging than the \textit{Tabletop} setting as it requires the grasping algorithm to avoid collisions with the bin while grasping.
    \myitem \textit{Wall.} Robot grasps from object hung on a flat wall 1m in front of the robot. 
    \myitem \textit{Random.} Robot grasps from a pile of objects randomly dumped into a bin that is randomly positioned in the workspace with a random height (0-15cm to tabletop) and random tilt angle (0-30$\degree$ to tabletop).
\end{itemize}
For each configuration, we run a total of 10 test runs, where each run consists of 10 (\textit{Wall}) or 20 (others) grasping episodes. Objects are replaced in the scene after each test run.
Each grasping episode begins with the robot's initial gripper positioned in a pose such that all target objects are visible to the wrist mounted camera.

Since the algorithm formulation predicts only relative 6DoF position, it works out-of-the-box with any initial starting position.
Row [pretrain only] in Tab. \ref{tab:setup} shows the same model trained with only human demonstration data without any fine-tuning on the real robot. We can see that this model is able to perform reasonably well  out-of-the-box across different scene configurations, due to the diversity of the demonstrations. 
Fine-tuning under each specific setting further improves the performance around $18\%$ on average ([+finetune] in Tab. \ref{tab:setup}).

\begin{figure}
    \centering
    \includegraphics[width=\linewidth]{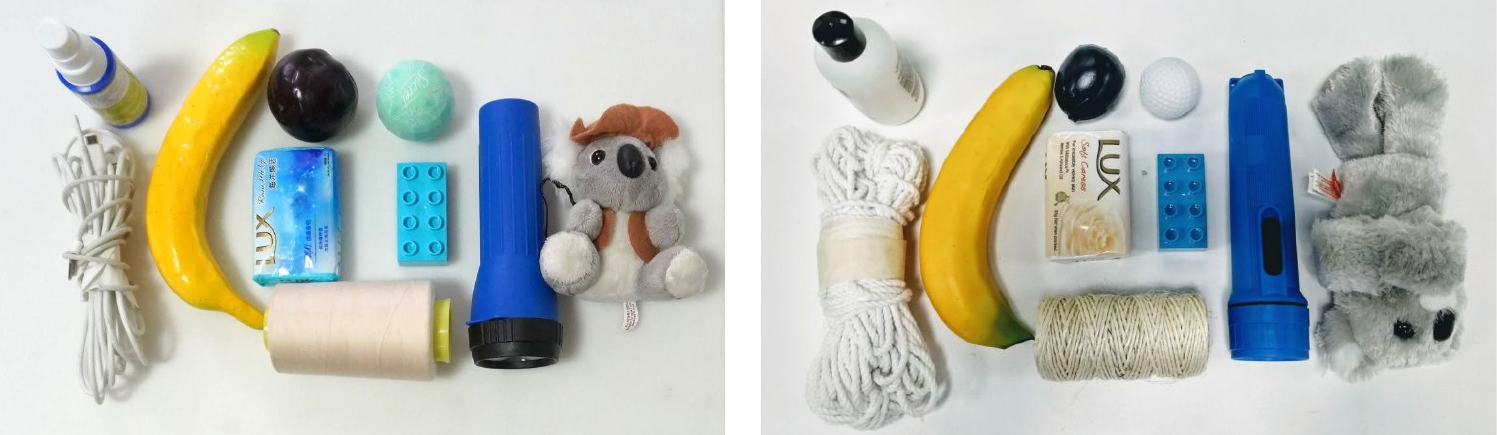}
    \caption{The testing objects (left) used to reproduce the dynamic grasping in clutter experiments of \cite{viereck2017learning,morrison2018closing} (right).}
    \label{fig:dy_object}
\vspace{2mm}
    \centering
    \includegraphics[width=\linewidth]{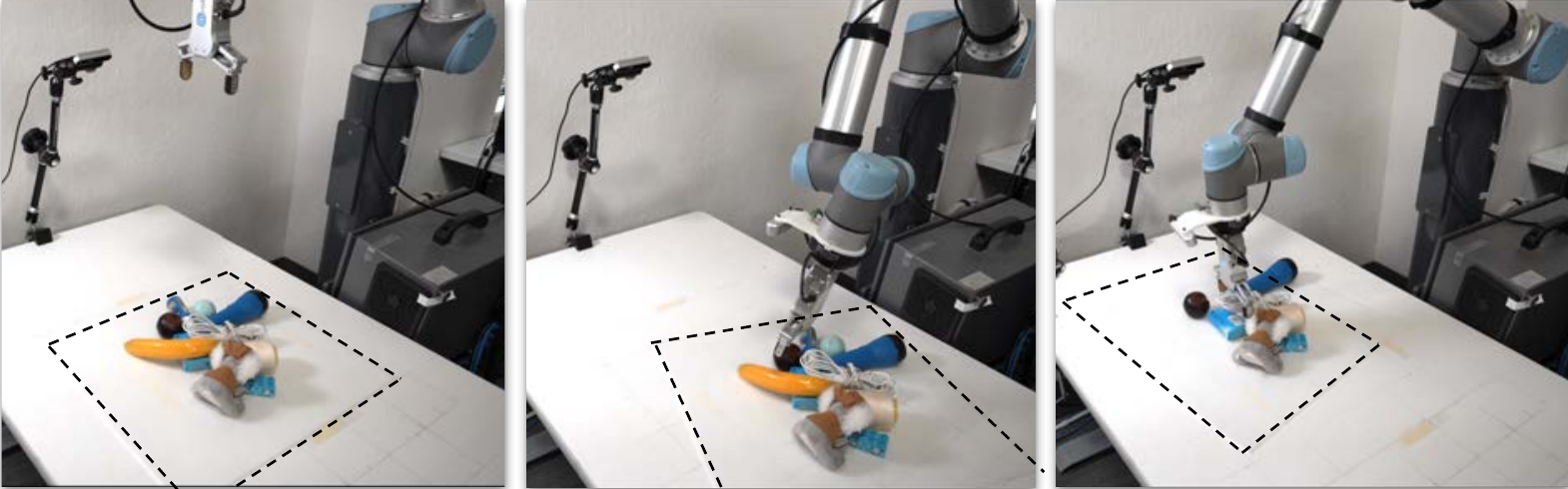}
    \caption{In dynamic scene experiments, the entire pile of objects is randomly shifted around while the gripper approaches an object.}
    \label{fig:dynamic}
    \vspace{-3mm}
\end{figure}

\mypara{Grasping in dynamic settings.}
We also test our algorithm's grasping performance in dynamic settings using the same experimental setup as Morrison \etal \cite{morrison2018closing}.
During each test run, we arrange a pile of 10 objects (Fig. \ref{fig:dy_object}) on a movable sheet on a tabletop. The robot attempts multiple grasps -- any objects that are grasped are removed.
During each grasping attempt (\ie episode), the pile is moved once by hand randomly (using the movable sheet). The movements have translations $>0.1$m and rotations $>25\degree$ (Fig. \ref{fig:dynamic}). 
This continues until all objects in the pile are grasped, or at least three consecutive grasps fail. 
We execute 10 test runs and average the grasping performance across the runs.
Tab. \ref{tab:dynamic} column [Dynamic] reports these results and their comparisons to alternative approaches in the same dynamic setting. 
These results show that our algorithm is able to achieve higher grasping success rates compared to alternative approaches for both static and dynamic settings.

\begin{table}[h]
\centering
\caption{Comparison to state-of-the-art methods (mean \%). \label{tab:dynamic} }
\vspace{-3mm}
\begin{tabular}{l ccc}
\toprule
Method and Setup  & Static Scenes & Dynamic Scenes & Time\\ \midrule
GG-CNN \cite{morrison2018closing}  & 87 $\pm$ 7    & 81 $\pm$ 8 &19ms    \\
Viereck \etal \cite{viereck2017learning} & 89 &　77 &　0.2s \\
Zeng \etal \cite{zeng2018robotic} & 90 $\pm$ 6 & - &- \\
Ours & 92$\pm$5 & 88 $\pm $ 8  & 0.18s \\ 
\bottomrule
\end{tabular}
\vspace{-2mm}
\end{table}

\begin{figure}
    \centering
    \includegraphics[width=0.9\linewidth]{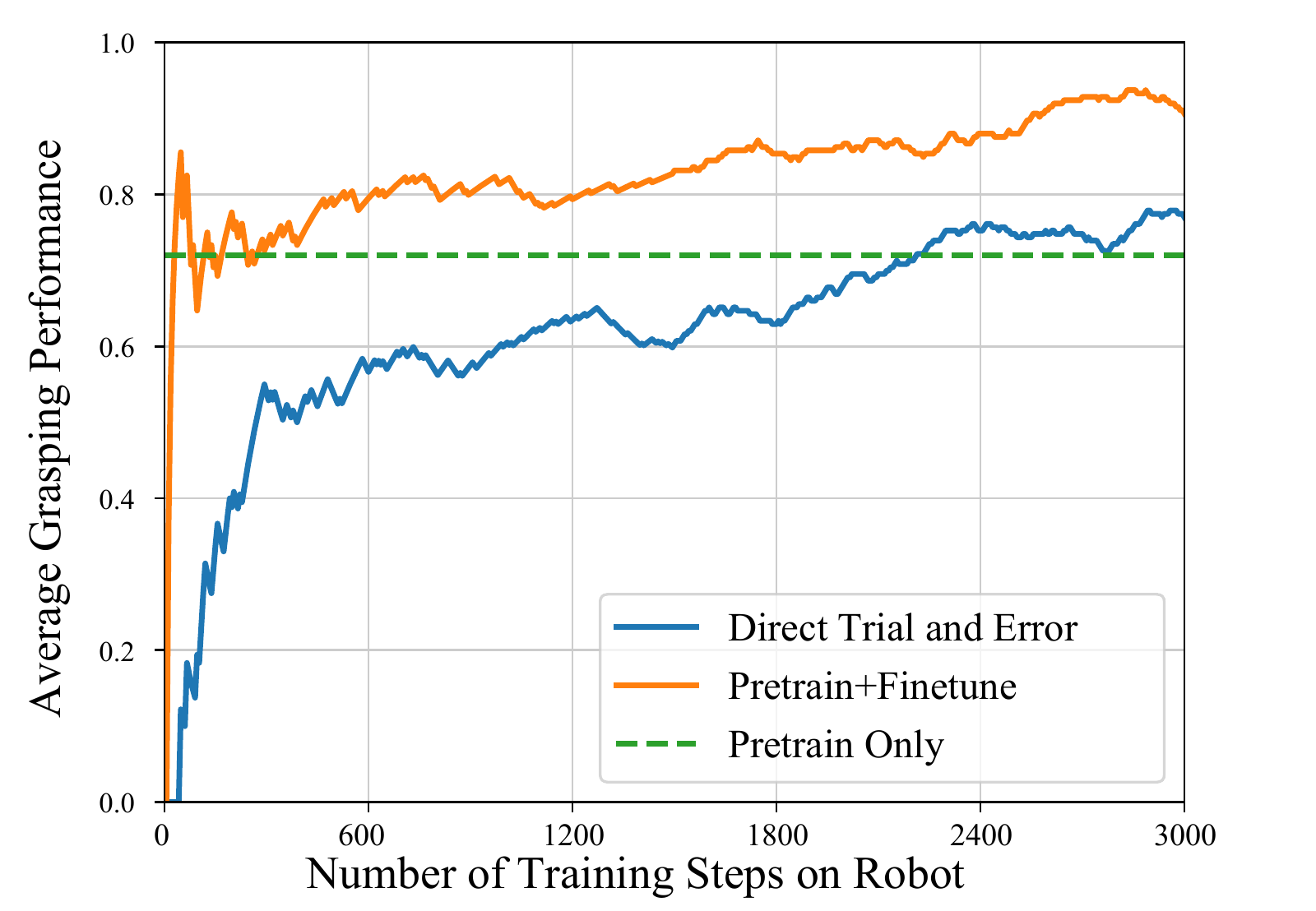}
     \vspace{-3mm}
    \caption{Grasping performance of our algorithm with and without pretaining on the demonstration data in the ``Tabletop'' setting.}
    \label{fig:effect_demo}
    \vspace{-3mm}
\end{figure}

\mypara{Effect of pretaining with demonstration data.}
To evaluate the benefits of pretraining on human demonstration data, we compare the our algorithm's performance with a model directly trained from on-robot self-supervised trial and error (described in Sec. \ref{sec:alg}). 
Fig. \ref{fig:effect_demo} plots grasping success vs. training iterations, where each iteration happens every five grasping episodes.
The diverse training data collected from human demonstrations not only helps the algorithm learn faster (higher performance in the early training stage), but also helps the algorithm learn better (higher performance after fine-tuning). 
This experiment shows that human demonstration data is more effective than trial and error data since the demonstration data contains significantly more diverse grasping examples than the trial and error data collected on the robot. This diversity is important for pretraining grasping policies that can generalize to different grasping scenarios.  

\section{Conclusions and Future Work} 
We introduce a new low-cost hardware interface for collecting grasping demonstrations in diverse environments, and a visual 6DoF closed-loop grasping algorithm that uses action-view based rendering. Our experiments demonstrate that training on the demonstration data improves both grasping performance and learning efficiency, and the capacity to move in 6DoF and adaptive closed-loop control enabled the algorithm to handle a variety of environments. 

Our system is not without limitations.
Our approach uses simple view-based rendering as a forward predictive model. While this approach can model possible motions and passive observations, it does not model the contact physics, which may be important during in-contact manipulation. It would be interesting to extend our predictive model with a learnable function that considers object and contact physics \cite{xu2019densephysnet}.
More broadly, view-based rendering may also be applicable for other tasks with ego-centric visual states and action spaces -- investigating its benefits for other applications (\eg navigation) would be interesting future work.
It would also be interesting to investigate how to make use of the other information captured in the demonstration (e.g., placing trajectories) for other applications (e.g., placing \cite{zakka2019form2fit}). 

\section*{Acknowledgment}
We would like to thank Stefan Welker and Ivan Krasin for their help on designing the handheld gripper, and Google Robotics for hardware and operational support.  We are also grateful for financial support from Amazon.

\bibliographystyle{IEEEtran}
\bibliography{main}


\begin{appendices}
\section{Data Collection Device: Hardware Details}
\begin{figure}[h]
    \centering
    \includegraphics[width=\linewidth]{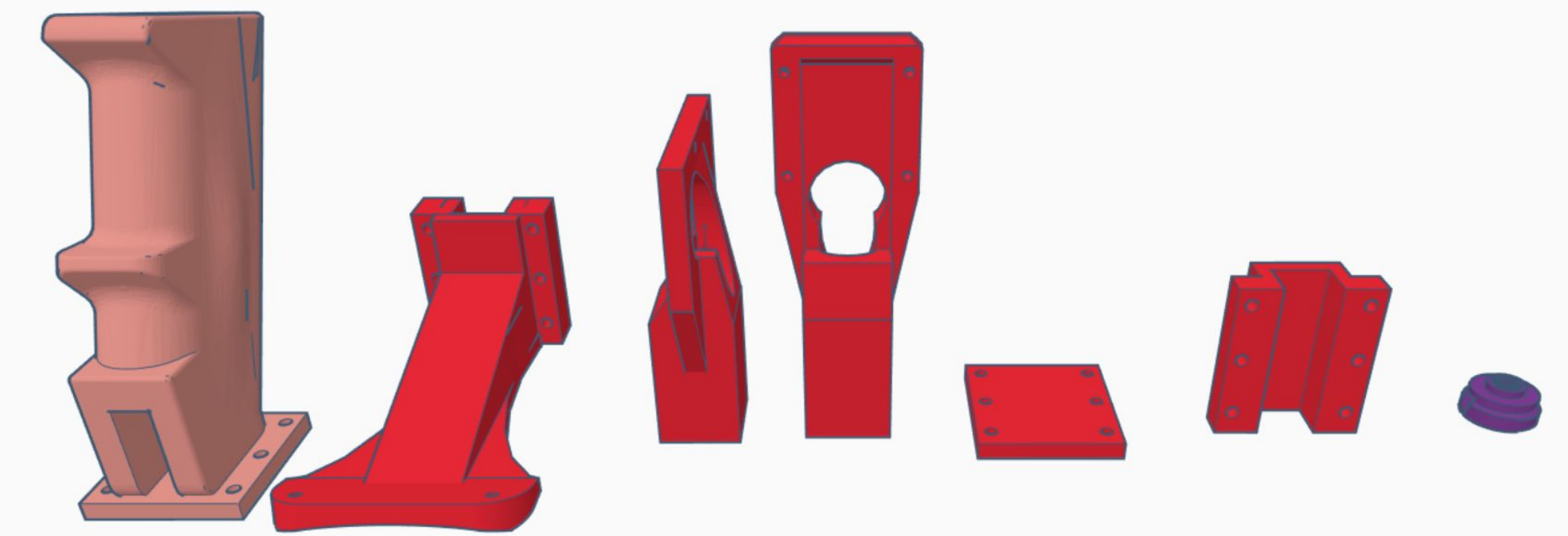}
    \caption{CAD models for 3D printed parts..}
    \label{fig:part}
\end{figure}

Table \ref{tab:part} provides a list of hardware components (and associated costs) used to build our handheld data collection device. 
Figure \ref{fig:part} shows CAD models for 3D printed parts, which can be download from our \href{http://graspinwild.cs.columbia.edu/}{project webpage}.

\begin{table}[h]
\centering
\setlength\tabcolsep{4px}
\begin{tabular}{lll}
\toprule
 Part Names                                      & Price (\$) &  \\
\midrule
 3D Printed Parts                                & 30    & - \\
Intel Compute Stick                             & 280    & \href{https://www.amazon.com/gp/product/B01DQ3RLPO/ref=oh_aui_detailpage_o02_s00?ie=UTF8&psc=1}{link} \\
Intel RealSense D415                            &   150  & \href{https://click.intel.com/intelr-realsensetm-depth-camera-d415.html}{link} \\
 Buck Converter 12V-\textgreater{}5V (5A)        & 3   &   
\href{https://www.amazon.com/dp/B078N1LRLR/ref=sspa_dk_detail_5?psc=1&pd_rd_i=B078N1LRLR&pf_rd_m=ATVPDKIKX0DER&pf_rd_p=f52e26da-1287-4616-824b-efc564ff75a4&pf_rd_r=Z9F5KPV1D4GCPHSMQ1M0&pd_rd_wg=qU8p1&pf_rd_s=desktop-dp-sims&pf_rd_t=40701&pd_rd_w=eoajw&pf_rd_i=desktop-dp-sims&pd_rd_r=40702ec5-c34a-11e8-a709-598374602c47}{link} \\
Battery 12V 6000mAh/5V 12000mAh       & 34  & \href{https://www.amazon.com/gp/product/B00ME3ZH7C/ref=oh_aui_detailpage_o02_s00?ie=UTF8&psc=1}{link}\\
 Monitor (1024x600 Touch Screen) &  60 & \href{https://www.amazon.com/GeeekPi-1024x600-Capacitive-Monitor-Raspberry/dp/B075QCXLPF/ref=sr_1_sc_2?s=electronics&ie=UTF8&qid=1538158550&sr=1-2-spell&keywords=7+inch+capative++LCD+HDMI}{link}\\
Dynamixel AX-12A Serial Servo             &    45 &\href{https://www.robotshop.com/en/dynamixel-ax-12a-smart-servo-serial.html}{link}  \\
 PP-Nest 12mm Push Button                        & 1   &\href{https://www.amazon.com/gp/product/B01LYZSEVB/ref=oh_aui_search_detailpage?ie=UTF8&psc=1}{link} \\

\midrule
Total Price                      & 603 \\
\bottomrule

\end{tabular}
\vspace{1mm}
\caption{Part list for our handheld data collection device. \label{tab:part}}
\end{table}

\section{Network Architecture}
The input to the current state encoder is a $640\times360$ RGB-D image and its corresponding surface normal map. The encoder uses the following network architecture (Conv2d represents one 2D convolution layer, ResBlock represent one residual block \cite{he2016deep} with BatchNorm \cite{ioffe2015batch}): \\

\noindent Conv2d(input=7, filter=64, kernel=3, stride=2, padding=1)\\
BatchNorm2d(64)\\
ReLU\\
ResBlock(input=64, filter=128,dilation=1)\\
MaxPool2d(kernel=3,stride=2,padding=1))\\
ResBlock(input=128, filter=128,dilation=1)\\
MaxPool2d(kernel=3,stride=2,padding=1))\\
ResBlock(input=128, filter=128,dilation=1)\\

The future state encoder uses the following: \\

\noindent Conv2d(input=7, filter=64, kernel=3, stride=2, padding=1)\\
BatchNorm2d(64)\\
ReLU\\
ResBlock(input=64, filter=128,dilation=1)\\
ResBlock(input=128, filter=128,dilation=1)\\
ResBlock(input=128, filter=128,dilation=1)\\

The action selection network uses the following: \\

\noindent ResBlock(input=256, filter=128,dilation=1)\\
ResBlock(input=128, filter=128,dilation=1)\\
ResBlock(input=128, filter=64,dilation=1)\\
Conv2d(input=64, filter=1, kernel=1, stride=1, padding=1)\\

\end{appendices}

\end{document}